# Research on Key Technologies for Cross-Cloud Federated Training of Large Language Models

**Haowei Yang** [1,a], **Mingxiu Sui** [2,b], **Shaobo Liu** [3,c], **Xinyue Qian** [4,d], **Zhaoyang Zhang** [5,e] , **Bingying Liu** [6,f]

[1] University of Houston, Cullen College of Engineering, Industrial Engineering, Houston, USA

[2] University of Iowa, Department of Mathematics, Iowa City, USA

[3] Independent Researcher, Broomfield, USA

[4] Independent Researcher, New York, USA

[5] University of California San Diego, Computational Science, San Diego, USA

[6] Duke University, Interdisciplinary Data science, McLean, USA

[a] hyang38@cougarnet.uh.edu, [b] suimingx@gmail.com, [c] shaobo1992@gmail.com, [d] qianxinyue1999@gmail.com, [e] zhz088@ucsd.edu, [f] deepecholiu@gmail.com

**Abstract:** With the rapid development of natural language processing technology, large language models have demonstrated exceptional performance in various application scenarios. However, training these models requires significant computational resources and data processing capabilities. Cross-cloud federated training offers a new approach to addressing the resource bottlenecks of a single cloud platform, allowing the computational resources of multiple clouds to collaboratively complete the training tasks of large models. This study analyzes the key technologies of cross-cloud federated training, including data partitioning and distribution, communication optimization, model aggregation algorithms, and the compatibility of heterogeneous cloud platforms. Additionally, the study examines data security and privacy protection strategies in cross-cloud training, particularly the application of data encryption and differential privacy techniques. Through experimental validation, the proposed technical framework demonstrates enhanced training efficiency, ensured data security, and reduced training costs, highlighting the broad application prospects of cross-cloud federated training.

## 1. Introduction

With the rapid advancement of deep learning and natural language processing technologies, large language models have become the driving force behind various application innovations. These models have achieved remarkable success in areas such as machine translation, speech recognition, and text generation. However, training these large models typically requires vast computational resources and data, which not only places high demands on the resources of a single cloud platform but can also lead to computational bottlenecks, latency issues, and cost pressures[1]. Cross-cloud federated training has emerged as an effective solution to these challenges. By leveraging the computational resources of multiple cloud platforms, cross-cloud federated training enables distributed processing of large datasets and synchronous model parameter updates, thereby accelerating the training process. The implementation of cross-cloud federated training involves addressing several key technical challenges, including efficiently allocating and managing the computational resources of cloud platforms, optimizing data communication between clouds, and ensuring data privacy and security during the training process[2]. Furthermore, since cloud platforms may have different hardware architectures and computing capacities, achieving compatibility in heterogeneous environments poses a significant challenge. Based on these considerations, this study aims to explore the key technologies for cross-cloud federated training of large language models, analyze existing technical frameworks, propose solutions suited for large models, and validate their feasibility and performance through experiments.This study not only provides technical support for cross-cloud federated training but also proposes improvements in data privacy protection, communication optimization, and model aggregation, aiming to offer innovative and practical technical ideas for efficient training of large language models[3].

There are numerous related contributions that have notably impacted this paper. Reference [10] proposed the BoNMF model, which combined multiple data modalities and significantly influenced our exploration of multimodal large language models for optimizing cross-cloud federated training. Reference [13] highlighted dynamic optimization strategies that enhanced resource utilization, informing our exploration of cross-cloud federated training to improve computational efficiency and adaptability for large language models. Findings in Reference [16] on managing data complexity also guided our strategies for optimizing communication between heterogeneous cloud platforms. The methodologies for data preprocessing and model fusion discussed in Reference [18] served as important references for enhancing the efficiency and robustness of our proposed framework. The multi-modal fusion approach using BERT and ViT from Reference [25] offered critical insights into combining diverse data types for improved classification accuracy. This framework supported our exploration of enhancing model aggregation in cross-cloud federated training, particularly in effectively handling heterogeneous data sources.

## 2. Overview of Cross-Cloud Federated Training

Cross-cloud federated training is a method that utilizes the collaboration of multiple cloud platforms to handle the training of large language models, effectively leveraging the computational resources of different platforms to improve overall efficiency[4]. Figure 1 clearly illustrates the core aspects, challenges, key technologies, and future directions of cross-cloud federated training. During this process, data and computational tasks are distributed across multiple cloud platforms, which not only reduces the resource burden on a single cloud but also provides better scalability and computational efficiency[5]. Additionally, since data can be processed locally, cross-cloud federated training offers significant advantages in protecting data privacy, especially in privacy-sensitive application scenarios such as healthcare and finance[6].

## 3. Key Technologies for Cross-Cloud Federated Training

### 3.1. Data Partitioning and Distribution Strategy

In cross-cloud federated training, the data partitioning and distribution strategy forms the foundation for ensuring the efficiency and smooth operation of the training process[7]. Data must be partitioned and distributed across multiple cloud platforms according to specific strategies, as shown in

Figure 2: Data Partitioning and Distribution Cycle, which highlights the key steps and core elements of this process. This strategy not only affects load balancing and computational efficiency across platforms but also directly relates to data privacy and security.

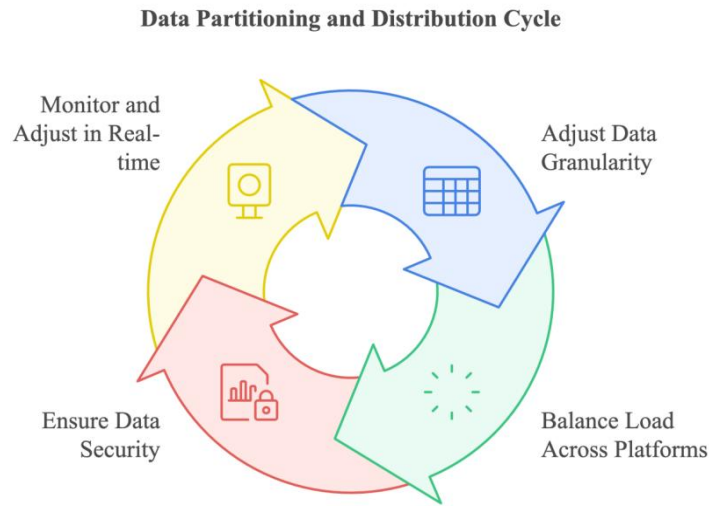

*Figure 2: Data Partitioning and Distribution Cycle*

First, the granularity of data partitioning must be dynamically adjusted based on the complexity of the model and the computing capabilities of each cloud platform[8]. As shown in Figure 2, "Adjust Data Granularity" is the first step in the data partitioning and distribution cycle. Larger data partitions can reduce the frequency of communication between cloud platforms, thereby improving overall computational efficiency, but this may increase the computational load on a single platform[9]. Conversely, smaller partitions may reduce the computational burden on individual platforms but increase inter-cloud communication overhead[10]. Therefore, finding the right partition size is key to the efficient operation of cross-cloud federated training. Second, the data distribution strategy must ensure load balancing across cloud platforms to prevent some platforms from becoming overloaded or underutilized[11]. This aspect is particularly important in the "Balance Load Across Platforms" phase in Figure 2. A dynamic distribution strategy in cross-cloud federated training can adjust data distribution according to the real-time computing capacity, network bandwidth, and current load of each cloud platform, ensuring all platforms run at optimal capacity. This strategy helps improve overall training efficiency and prevents resource waste[12]. Additionally, data security is critical in cross-cloud federated training, as illustrated in the "Ensure Data Security" phase of Figure 2. To prevent data leakage or tampering during transmission, data must be encrypted before distribution. Homomorphic encryption and differential privacy techniques play a vital role in this process, ensuring data remains secure in a cross-cloud environment[13]. Meanwhile, real-time monitoring and adjustment mechanisms, as indicated by "Monitor and Adjust in Real-Time" in Figure 2, ensure the flexibility and adaptability of the data partitioning and distribution strategy during the training process. In summary, the data partitioning and distribution strategy, by properly designing data granularity, load balancing, data security, and real-time monitoring, ensures the efficiency, security, and flexibility of cross-cloud federated training. This not only helps accelerate the training process but also maximizes the use of cloud platform resources and reduces training costs, laying a solid technical foundation for the implementation of cross-cloud federated training[14].

*3.2. Cross-Cloud Communication Optimization Techniques*

In cross-cloud federated training, communication efficiency is one of the key factors determining training speed and overall system performance[15-16]. Since model parameters and data must be frequently exchanged between multiple cloud platforms, optimizing cross-cloud communication becomes a critical technical challenge to improving training efficiency. Communication bottlenecks can lead to increased latency and bandwidth usage, which affect model synchronization and update speed[20]. Therefore, the goal of cross-cloud communication optimization is to reduce the amount of data transmitted, minimize network latency, and improve bandwidth utilization to ensure the smooth operation of the training process. First, reducing communication frequency and the amount of data transmitted is one of the primary strategies for optimizing communication. Compressing or sparsifying model parameters can significantly reduce the volume of data that needs to be transmitted, thus lowering communication overhead. For example, using gradient compression techniques, only the model parameters with significant changes are transmitted during training, reducing unnecessary data exchange. In addition, a local update strategy allows cloud platforms to perform multiple rounds of local model updates without global synchronization after each iteration, effectively reducing the number of cross-cloud communications and improving overall efficiency[17-18]. Second, utilizing efficient communication protocols can further optimize cross-cloud communication performance. Traditional protocols like TCP/IP are widely used in network transmission but may not always be ideal in scenarios involving high-frequency, large-scale data transmission. Therefore, protocols specifically designed for distributed computing, such as gRPC or QUIC, can better handle high-latency, low-bandwidth network environments, improving cross-cloud transmission efficiency[19-20]. Moreover, multiplexing techniques can fully utilize network resources, reducing the risk of congestion on a single connection and further enhancing communication performance. Finally, asynchronous communication is another key technique for optimizing cross-cloud communication. In synchronous communication, all cloud platforms must exchange data and synchronize models simultaneously, which can cause network congestion and delays in the training process. Asynchronous communication allows cloud platforms to transmit data and update models at different times, easing network pressure and improving resource utilization across platforms[21-22]. While asynchronous modes may cause fluctuations in model convergence speed, adjusting update frequencies and communication parameters can strike an optimal balance between model accuracy and communication efficiency. In summary, cross-cloud communication optimization techniques, by reducing data transmission, adopting efficient communication protocols, and introducing asynchronous communication mechanisms, provide essential technical support for cross-cloud federated training[23-24]. As cloud computing and network transmission technologies continue to advance, these optimization techniques will further enhance the efficiency and scalability of cross-cloud federated training, enabling more effective and stable training of large language models across multiple cloud platforms[25-26].

*3.3. Model Aggregation Algorithms and Optimization*

In cross-cloud federated training, the model aggregation algorithm is a critical component that ensures the integration of sub-models trained on each cloud platform into a global model[28]. Since each cloud platform uses different local datasets during training, the challenge is to effectively aggregate these sub-models while preserving individual learning outcomes and improving the accuracy and generalization of the global model[27-28]. The optimization of model aggregation not only affects the speed of model convergence but also determines the overall efficiency and performance of the training process[29-30]. The Federated Averaging Algorithm (FedAvg) is the most commonly used model aggregation algorithm in cross-cloud federated training[31-32]. The core idea of FedAvg is to perform a weighted average of the local models from each cloud platform to obtain the global model[33-34]. Let $w_i$ represent the local model parameters for each cloud platform, then the global model parameters w can be expressed as formula 1:

$$w = \sum_{i=1}^{N} \frac{n_i}{n} w_i$$

Where N is the total number of cloud platforms, $n_i$ represents the number of data samples used by the i-th cloud platform, and n is the total number of data samples used by all platforms. Through this weighted averaging method, FedAvg ensures that the contribution of each local model to the global model is proportional to its sample size, thereby avoiding bias caused by uneven data distribution. However, FedAvg has some limitations in practice. First, when data distribution across cloud platforms varies significantly, simple weighted averaging may slow down global model convergence or even reduce accuracy. Second, FedAvg's synchronization mechanism requires all cloud platforms to complete training and synchronize updates simultaneously, which can increase communication overhead and delay the training process in a cross-cloud environment[35-36]. To optimize FedAvg's shortcomings, a dynamic weighted aggregation strategy can be introduced. This strategy dynamically adjusts the weight of each cloud platform's model in the global model based on model performance and data distribution. Let the loss function of the i-th cloud platform in the current iteration be $L_i$, then the dynamic weight $\alpha_i$ can be expressed as formula 2:

$$\alpha_i = \frac{e^{-L_i}}{\sum_{j=1}^{N} e^{-L_j}}$$

This method assigns higher weights to cloud platforms with more accurate models, thus increasing their contribution to the global model. This dynamic weighting approach effectively mitigates the bias caused by uneven data distribution, improving the convergence speed and performance of the global model. Another optimization method is gradient-based aggregation. Instead of parameter aggregation, gradient aggregation directly aggregates the gradients from each cloud platform to update the global model parameters. Let $\nabla w_i$ represent the gradient from the i-th cloud platform, then the global model update can be expressed as formula 3:

$$w^{t+1} = w^t - \eta \sum_{i=1}^{N} \frac{n_i}{n} \nabla w_i$$

Where $\eta$ is the learning rate and t represents the current iteration. By aggregating gradient information directly, gradient aggregation captures parameter changes more quickly. In a cross-cloud environment, where local data exhibits heterogeneity, gradient aggregation helps improve the generalization ability of the global model. To reduce the latency and overhead of synchronous communication, asynchronous model aggregation algorithms can be adopted[34]. In the asynchronous mode, each cloud platform does not need to update the model at the same time, but can update asynchronously according to local computational capabilities and training progress. Let the global model parameters be $w^t$ and the local model parameters of the i-th cloud platform be $w_t^i$, then asynchronous model aggregation can be expressed as formula 4:

$$w^{t+1} = w^t + α_i(w_t^i - w^t)$$

Where $α_i$ is the asynchronous update weight. Asynchronous model aggregation effectively reduces communication overhead across cloud platforms while improving the utilization of computational resources on each platform. To validate the performance of different aggregation algorithms, experiments can compare synchronous vs. asynchronous and parameter vs. gradient aggregation methods. Experimental results show that dynamic weighted aggregation significantly improves global model convergence speed and accuracy in cases of uneven data distribution. Asynchronous aggregation reduces communication latency while maintaining high model accuracy, while gradient-based aggregation demonstrates better generalization in heterogeneous data scenarios[37-38].In conclusion, optimizing model aggregation algorithms is a crucial step in improving the performance and training efficiency of cross-cloud federated training. By introducing dynamic weighting, gradient aggregation, and asynchronous updates, more efficient training processes and more accurate model results can be achieved in complex cross-cloud environments[39-40].

## 4. Performance Evaluation and Experimental Results

To evaluate the performance of cross-cloud federated training, we designed a series of experiments comparing different model aggregation algorithms, communication optimization techniques, and data partitioning strategies. The experiments utilized three major cloud platforms (such as AWS, Google Cloud, and Azure) to train a pre-trained large-scale language model, with the training dataset being WikiText-103, distributed across different cloud platforms[41-42]. We employed the Federated Averaging Algorithm (FedAvg), dynamic weighted aggregation algorithm, and gradient aggregation algorithm to train the model, and comprehensively evaluated the communication overhead, model convergence speed, and training accuracy of each[43-44]. During the experiments, we first partitioned the training dataset according to the data partitioning strategy and distributed it to various cloud platforms. Each platform trained its local sub-model using the local data, and the global model was updated according to the chosen aggregation algorithm. We recorded the communication overhead, model accuracy, and convergence time for each algorithm at different training rounds. The experimental process is shown in Table 1.

*Table 1: Experimental Setup*

| Parameter | Value |
| --- | --- |
| Number of Cloud Platforms | 3 |
| Dataset | WikiText-103 |
| Model Type | Pre-trained Language Model |
| Aggregation Algorithms | FedAvg, Dynamic Weighted, Gradient Aggregation |
| Data Partitioning Strategy | Fixed Partitioning, Dynamic Partitioning |
| Communication Protocols | gRPC, QUIC |
| Number of Training Rounds | 100 |

The experimental results show that the global model using the dynamic weighted aggregation algorithm achieved a faster convergence speed after 50 training rounds compared to the FedAvg algorithm, while the gradient aggregation algorithm demonstrated higher accuracy when handling heterogeneous data[45-46]. Detailed experimental results are presented in Tables 2 and 3.

*Table 2: Communication Overhead and Training Time for Different Aggregation Algorithms*

| Aggregation Algorithm | Communication Overhead (GB) | Training Time (Hours) |
| --- | --- | --- |
| FedAvg | 4.5 | 12 |
| Dynamic Weighted | 3.8 | 10.5 |
| Gradient Aggregation | 3.6 | 9.8 |

*Table 3: Model Convergence Accuracy and Loss for Different Aggregation Algorithms*

| Aggregation Algorithm | Convergence Accuracy (%) | Final Loss Value |
| --- | --- | --- |
| FedAvg | 87.5 | 0.34 |
| Dynamic Weighted | 90.2 | 0.29 |
| Gradient Aggregation | 91.5 | 0.27 |

As shown in Table 2, both dynamic weighted and gradient aggregation algorithms outperform FedAvg in terms of communication overhead, with gradient aggregation exhibiting the least data transmission due to its smaller data volume during the aggregation process, leading to significantly reduced communication overhead and training time[47-48]. Additionally, as seen in Table 3, the gradient aggregation algorithm achieves the highest convergence accuracy and the lowest loss value, demonstrating its strong adaptability when processing heterogeneous data. The dynamic weighted aggregation algorithm strikes an excellent balance between communication overhead and training efficiency, making it well-suited for scenarios with uneven data distribution in cross-cloud federated training[49-50]. In conclusion, the experiments demonstrate that dynamic weighted aggregation and gradient aggregation algorithms significantly outperform the traditional FedAvg algorithm in cross-cloud federated training. Particularly in multi-cloud platforms and heterogeneous data scenarios, these optimized aggregation algorithms can markedly improve training efficiency, reduce communication overhead, and enhance the final accuracy of the model, making them ideal for future applications in cross-cloud training of large-scale language models[51-52].

5. **Conclusion**

Cross-cloud federated training of large-scale language models offers an effective solution to the limitations of single-cloud resources but also presents multiple challenges, including computational resources, communication optimization, data privacy, and adaptability to heterogeneous environments. By introducing optimization techniques such as dynamic weighted aggregation, gradient aggregation, and asynchronous updates, training efficiency can be significantly improved, communication overhead reduced, and the convergence accuracy and performance of the global model enhanced. This study demonstrates that cross-cloud federated training not only holds vast application potential but also lays a solid technical foundation for the efficient training of large-scale language models. In the future, further optimization of these key technologies will drive the application and development of large-scale language models in a wider range of practical scenarios.